\documentclass[conference]{IEEEtran}
\IEEEoverridecommandlockouts

\usepackage[T1]{fontenc}
\usepackage[utf8]{inputenc}
\usepackage{amsmath,amssymb,amsfonts}
\usepackage{graphicx}
\usepackage{booktabs}
\usepackage{caption}
\usepackage{subcaption}
\usepackage{array}
\usepackage{float}
\usepackage{algorithm}
\usepackage{algpseudocode}
\usepackage{multirow}
\usepackage{textcomp}
\usepackage{xcolor}
\usepackage{cite}
\usepackage[colorlinks=true,linkcolor=blue,citecolor=blue]{hyperref}

\def\BibTeX{{\rm B\kern-.05em{\sc i\kern-.025em b}\kern-.08em
    T\kern-.1667em\lower.7ex\hbox{E}\kern-.125emX}}

\begin{document}

\title{Adversarial Vulnerability Transcends Computational Paradigms: Feature Engineering Provides No Defense Against Neural Adversarial Transfer}

\author{
\IEEEauthorblockN{Achraf Hsain\IEEEauthorrefmark{1}, Ahmed Abdelkader\IEEEauthorrefmark{1}, Emmanuel Baldwin Mbaya\IEEEauthorrefmark{2}, Hamoud Aljamaan\IEEEauthorrefmark{1}\IEEEauthorrefmark{3}}
\IEEEauthorblockA{\IEEEauthorrefmark{1}Department of Information \& Computer Science, KFUPM, Dhahran, Saudi Arabia}
\IEEEauthorblockA{\IEEEauthorrefmark{2}Department of Computer Engineering, KFUPM, Dhahran, Saudi Arabia}
\IEEEauthorblockA{\IEEEauthorrefmark{3}Interdisciplinary Research Center for Finance and Digital Economy, KFUPM, Dhahran, Saudi Arabia}
}

\maketitle

\begin{abstract}
Deep neural networks are vulnerable to adversarial examples---inputs with imperceptible perturbations that cause confident misclassifications. While adversarial transferability within neural network families is extensively documented, a fundamental question remains unexplored: do classical machine learning pipelines operating on handcrafted features inherit this vulnerability when attacked via perturbations crafted on neural surrogates? Feature engineering creates an explicit information bottleneck through gradient quantization and spatial binning, suggesting a plausible defense mechanism that could filter high-frequency adversarial signals. We systematically evaluate this hypothesis through the first comprehensive study of adversarial transfer from deep neural networks to Histogram of Oriented Gradients (HOG) based classifiers. Using VGG16 as a surrogate, we generate FGSM and PGD adversarial examples and evaluate their transfer to four classical classifiers (K-Nearest Neighbors, Decision Tree, Linear SVM, Kernel SVM) and a shallow neural network across eight HOG configurations on CIFAR-10. Our results strongly refute the protective hypothesis: all classifiers suffer substantial accuracy degradation ranging from 16.6\% to 59.1\% relative drop, comparable to or exceeding neural-to-neural transfer baselines. More surprisingly, we discover \textit{attack hierarchy reversal}---contrary to established patterns where iterative PGD dominates single-step FGSM within neural networks, FGSM causes greater degradation than PGD in 100\% of classical ML cases, suggesting iterative attacks overfit to surrogate-specific features that do not survive feature extraction. We further characterize HOG parameter sensitivity, finding that block normalization provides partial but insufficient mitigation. These findings demonstrate that adversarial vulnerability is not an artifact of end-to-end differentiability but a fundamental property of image classification systems, with significant implications for security-critical deployments across computational paradigms.
\end{abstract}


\section{Introduction}

Deep neural networks have achieved remarkable success across computer vision tasks, yet they remain vulnerable to \textit{adversarial examples}---inputs modified with carefully crafted, often imperceptible perturbations that cause confident misclassifications~\cite{fgsm,pgd}. A particularly concerning property of these attacks is \textit{transferability}: adversarial examples generated against one model frequently fool entirely different models, enabling practical black-box attacks without knowledge of the target system's architecture or parameters~\cite{fgsm}.

The adversarial machine learning literature has extensively characterized transferability within neural network families---from CNNs to Vision Transformers, across different architectures and training regimes~\cite{mahmood2021,waseda2023}. However, a fundamental question remains largely unexplored: \textit{do classical, feature-engineered classification pipelines share this vulnerability when attacked via transferred perturbations from neural surrogates?} This question carries significant practical implications, as many deployed systems---particularly in resource-constrained environments, legacy infrastructures, and interpretability-critical applications---continue to rely on traditional machine learning approaches operating on hand-crafted features.

The Histogram of Oriented Gradients (HOG) descriptor~\cite{hog} presents a particularly interesting case study. Unlike neural networks that learn hierarchical features end-to-end, HOG explicitly encodes local edge orientation distributions through gradient computation and spatial binning---a fundamentally different representation strategy. One might hypothesize that this coarse quantization process could attenuate high-frequency adversarial perturbations, providing natural robustness through the information bottleneck of feature engineering.

Recent work has begun exploring HOG-based systems in adversarial contexts, though from perspectives distinct from cross-paradigm transfer. Siddique et al.~\cite{siddique2020} proposed a lateralized learning architecture combining HOG and SIFT features for improved robustness, demonstrating that heterogeneous feature representations can enhance resilience against direct attacks. Mohammed et al.~\cite{mohammed2025} evaluated HOG-based classifiers for orthopedic disease detection under multiple adversarial attacks, finding substantial vulnerability when attacks target the HOG pipeline directly. Lal et al.~\cite{lal2021} proposed a defense framework combining HOG with other handcrafted and deep features alongside adversarial training, achieving high robustness against FGSM and DeepFool attacks. However, these studies either propose novel defense architectures, evaluate direct attacks on HOG systems, or combine HOG with adversarial training and feature fusion---none systematically investigates whether gradient-based adversarial perturbations \textit{transfer} from neural networks to HOG-based classifiers operating in isolation.

Prior work has also demonstrated that classical models can be compromised via learned black-box attacks on raw pixels~\cite{hayes2017}, but whether standard gradient-based attacks crafted on CNN surrogates survive HOG's explicit gradient quantization and spatial binning remains systematically unevaluated. This gap is particularly significant given the widespread deployment of HOG-based systems in security-critical applications where understanding cross-paradigm attack vectors is essential.

This paper addresses this gap through a rigorous empirical investigation of adversarial transferability from CNN surrogates to HOG-based classification pipelines. Specifically, we make the following contributions:

\begin{enumerate}
    \item We provide the \textbf{first systematic evaluation} of $L_\infty$-bounded adversarial transfer from deep neural networks to HOG-based classifiers, testing four classical ML models and a shallow neural network across eight HOG configurations---extending beyond prior work that examined only direct attacks or fixed HOG parameterizations.
    
    \item We discover and characterize \textbf{attack hierarchy reversal} in cross-paradigm transfer: contrary to established neural network patterns where iterative PGD dominates single-step FGSM, we observe FGSM causing greater degradation in 100\% of classical ML cases---a phenomenon not previously reported in the literature.
    
    \item We provide \textbf{systematic HOG parameter sensitivity analysis}, identifying which configuration choices (cell size, orientation bins, block normalization) most influence adversarial robustness---the first such analysis in the context of transferred perturbations.
    
    \item We quantify \textbf{perturbation budget effects} on cross-paradigm transfer, revealing nonlinear vulnerability scaling that differs from neural network behavior.
\end{enumerate}

Our findings challenge prevailing assumptions about the relationship between feature engineering and adversarial robustness, with significant implications for security-critical deployments of classical machine learning systems.

\section{Literature Review}

Adversarial examples---inputs modified with small, often imperceptible perturbations to cause misclassification---represent a foundational challenge in machine learning security. A key property enabling practical attacks is \textit{transferability}: adversarial examples crafted for one model often fool others without knowledge of the target's architecture. While neural-to-neural transferability is well-documented, its effect on classical, feature-engineered pipelines remains critically underexplored.

\subsection{Foundational Attacks}

Goodfellow et al.\ introduced the \textit{linearity hypothesis}, positing that adversarial vulnerability arises from locally linear model behavior in high-dimensional spaces~\cite{fgsm}. This insight motivated the \textbf{Fast Gradient Sign Method (FGSM)}, which crafts perturbations by moving each pixel in the direction of the loss gradient's sign, providing the first strong evidence of cross-model transferability~\cite{fgsm}.

Madry et al.\ framed robustness as min-max optimization and introduced \textbf{Projected Gradient Descent (PGD)}---a multi-step iterative attack that serves as the standard for evaluating and training robust models~\cite{pgd}. Together, these works established the core concepts of gradient-based attacks, transferability, and adversarial training.

\subsection{Enhancing Transferability}

Subsequent research has pursued two complementary strategies to improve transfer success: manipulating features in semantically meaningful spaces and shaping gradient computation.

\textbf{Feature-Space Methods.} Agarwal et al.\ demonstrated that frequency-domain manipulation via Redundant Discrete Wavelet Transform dramatically improves black-box transfer by targeting shared edge and texture features~\cite{agarwal2022}. Yang et al.'s \textbf{PatchAttack} uses reinforcement learning to place adversarial texture patches~\cite{yang2020}, while Gao et al.'s \textbf{Transferable Attentive Attack} targets discriminative regions via attention-guided losses~\cite{gao2021}. These methods converge on a key principle: perturbations exploiting fundamental visual features generalize better than those exploiting model-specific boundaries.

\textbf{Gradient-Shaping Methods.} Guo et al.'s \textbf{LinBP} improves transferability by skipping nonlinear activations during backpropagation~\cite{guo2020}. He et al.'s \textbf{Transformed Gradient} method amplifies and truncates gradients~\cite{he2022}. For targeted transfer, Weng et al.\ combine logit calibration with SVD-based feature mixing~\cite{weng2025}. These approaches modify \textit{how} perturbations are computed rather than \textit{what} is perturbed.

\subsection{Cross-Architecture and Cross-Paradigm Studies}

Mahmood et al.\ found that adversarial examples transfer strongly within architectural families but weakly across families (e.g., CNN to Vision Transformer)~\cite{mahmood2021}. Waseda et al.\ showed that transferred examples may induce different misclassification patterns across models~\cite{waseda2023}.

Crucially, Hayes \& Danezis demonstrated that classical models---Random Forests, SVMs, and k-NN---can be compromised via learned black-box attacks using only output scores~\cite{hayes2017}, establishing that non-differentiability is not a guaranteed defense. Maria et al.\ showed that HOG+SVM pipelines are sensitive to natural transformations~\cite{maria2019}, but no study has systematically evaluated gradient-based adversarial transfer to HOG features.

\subsection{HOG Features in Adversarial Contexts}

Recent work has begun exploring HOG-based systems under adversarial conditions, though from perspectives distinct from cross-paradigm transfer.

\textbf{Heterogeneous Feature Architectures.} Siddique et al.~\cite{siddique2020} proposed a lateralized learning system inspired by biological neural organization, combining HOG and SIFT features to process images at multiple levels of abstraction. Their system outperformed standard deep networks against FGSM attacks (81.06\% vs.\ 70.38\% for VGG under strong perturbations), demonstrating that heterogeneous feature representations can enhance robustness. However, their work proposes a novel architecture rather than evaluating transfer vulnerability, uses binary classification (cats vs.\ dogs) rather than multi-class benchmarks, and does not systematically vary HOG parameters.

\textbf{Direct Attacks on HOG Pipelines.} Mohammed et al.~\cite{mohammed2025} conducted a comprehensive evaluation of nine machine learning models using HOG features for orthopedic disease detection, testing robustness against FGSM, Basic Iterative Method (BIM), Label Poisoning Method (LPM), and Model Extraction Method (MEM) attacks. Their results revealed substantial vulnerability in HOG-based classifiers when attacks target the pipeline directly. Critically, their study generates attacks \textit{on} the HOG-based models themselves rather than investigating whether perturbations crafted on neural surrogates transfer to HOG representations---a fundamentally different threat model with distinct practical implications for heterogeneous system security.

\textbf{Feature Fusion Defenses.} Lal et al.~\cite{lal2021} proposed a defense framework combining HOG, LBP, and SFTA handcrafted features with DarkNet-53 deep features, along with adversarial training, for diabetic retinopathy classification under FGSM and DeepFool attacks. Their feature fusion approach achieved 99.9\% accuracy against adversarial examples. However, this robustness stems from the combination of adversarial training and multi-feature fusion---whether HOG features alone provide any inherent protection against transferred attacks was not evaluated. Moreover, their attacks were crafted directly on the target system rather than transferred from a different architecture.

\subsection{Research Gap and Our Contribution}

Despite these advances, a critical gap remains: no systematic evaluation exists for adversarial transfer from neural surrogates to HOG-based classifiers using standard $L_p$-bounded attacks. Prior work either proposes novel defense architectures~\cite{siddique2020}, evaluates direct attacks on HOG systems~\cite{mohammed2025}, demonstrates robustness through feature fusion and adversarial training without isolating HOG's contribution~\cite{lal2021}, or uses learned black-box attacks on raw pixels~\cite{hayes2017}. The HOG descriptor's explicit gradient quantization and spatial binning represent a fundamentally different computational paradigm from neural feature learning. Whether perturbations optimized against CNN loss landscapes survive this hand-crafted transformation---or whether the information bottleneck of feature engineering provides inherent protection---remains an open question.

This work directly addresses this gap by systematically evaluating FGSM and PGD adversarial transfer from a CNN surrogate (VGG16) to HOG-based classifiers across multiple configurations. Unlike prior work, we: (1) focus specifically on cross-paradigm transfer rather than direct attacks, (2) systematically vary HOG parameters to characterize sensitivity, (3) discover the attack hierarchy reversal phenomenon where FGSM outperforms PGD in transfer effectiveness, and (4) provide quantitative comparison against neural-to-neural transfer baselines.

Table~\ref{tab:comprehensive_comparison} synthesizes key methodologies and findings across the literature, positioning our contributions within this landscape.

\begin{table*}[!t]
\centering
\caption{Comparison of Adversarial Transferability Methods and Cross-Paradigm Findings}
\label{tab:comprehensive_comparison}
\renewcommand{\arraystretch}{1.3}
\scriptsize
\begin{tabular}{|p{1.8cm}|p{0.9cm}|p{2.2cm}|p{1.6cm}|p{2.2cm}|p{2.0cm}|p{2.4cm}|}
\hline
\textbf{Method} & \textbf{Category} & \textbf{Approach} & \textbf{Targets} & \textbf{Key Finding} & \textbf{Gap} & \textbf{Our Contribution} \\
\hline
\hline
\multicolumn{7}{|c|}{\textbf{Foundational Attacks}} \\
\hline
FGSM~\cite{fgsm} & Attack & Single-step gradient sign & DNNs & Demonstrates cross-model transfer via linearity & One-step only; neural targets & First evaluation on HOG classifiers; reveals stronger cross-paradigm transfer than PGD \\
\hline
PGD~\cite{pgd} & Attack & Multi-step iterative optimization & DNNs & Stronger than FGSM; robustness benchmark & May overfit to surrogate & Shows weaker HOG transfer than FGSM; supports overfitting hypothesis \\
\hline
\hline
\multicolumn{7}{|c|}{\textbf{Transfer Enhancement Methods}} \\
\hline
RDWT+PGD~\cite{agarwal2022} & Feature & Frequency-domain edge manipulation & CNNs & Edge features transfer across architectures & No classical ML testing & HOG edge encoding does not filter adversarial signal \\
\hline
LinBP~\cite{guo2020} & Gradient & Skip nonlinearities in backprop & DNNs & Linear gradients generalize better & Neural targets only & Standard FGSM sufficient for HOG transfer \\
\hline
NLC+TFM~\cite{weng2025} & Gradient & Logit calibration + SVD mixing & DNNs (targeted) & Removes model-specific features & Complex; targeted only & Untargeted attacks achieve significant HOG degradation \\
\hline
\hline
\multicolumn{7}{|c|}{\textbf{HOG-Specific Studies}} \\
\hline
Siddique~\cite{siddique2020} & Defense & Lateralized HOG+SIFT architecture & Binary class & Heterogeneous features improve robustness & Novel architecture; no transfer analysis & We evaluate existing pipelines under transfer \\
\hline
Mohammed~\cite{mohammed2025} & Analysis & HOG for orthopedic detection & ML models & HOG classifiers vulnerable to direct attacks & Direct attacks only; fixed HOG & We study CNN$\to$HOG transfer; vary parameters \\
\hline
Lal~\cite{lal2021} & Defense & HOG+LBP+SFTA+Deep fusion + AT & DR images & Feature fusion + AT achieves 99.9\% & Direct attacks; fusion required & We isolate HOG alone; show fusion/AT needed \\
\hline
\hline
\multicolumn{7}{|c|}{\textbf{Cross-Paradigm Studies}} \\
\hline
Mahmood~\cite{mahmood2021} & Analysis & CNN $\leftrightarrow$ ViT transfer & DNNs & Weak inter-family transfer & No classical ML & We show substantial CNN$\to$HOG transfer \\
\hline
Hayes~\cite{hayes2017} & Attack & Score-based learned perturbations & RF, SVM, k-NN & Classical ML vulnerable to black-box & Raw pixels; no HOG features & First HOG-specific transfer evaluation \\
\hline
Maria~\cite{maria2019} & Analysis & Natural transforms (blur, rotation) & HOG+SVM & HOG sensitive to edge corruption & No adversarial attacks & Adversarial perturbations cause greater degradation \\
\hline
\end{tabular}
\end{table*}

\section{Experimental Methodology}

This section details the experimental pipeline used to evaluate adversarial transferability from deep neural network surrogates to classical machine learning classifiers operating on HOG features. The methodology comprises five phases: dataset preparation and adversarial example generation (Sections~\ref{sec:dataset}--\ref{sec:adversarial_gen}), HOG feature extraction (Section~\ref{sec:hog_extraction}), hyperparameter optimization (Section~\ref{sec:hyperparameter}), neural network baselines (Section~\ref{sec:nn_baselines}), and the experimental evaluation protocol (Section~\ref{sec:protocol}).

\subsection{Dataset and Preprocessing}
\label{sec:dataset}

All experiments used the CIFAR-10 dataset~\cite{cifar10}, consisting of 60,000 color images ($32 \times 32 \times 3$) across 10 classes (50,000 training, 10,000 test). Images were upsampled to $64 \times 64$ pixels via bilinear interpolation to serve as the canonical space for both adversarial crafting and HOG extraction.

For neural network compatibility, a differentiable resolution bridge mapped $64 \to 224$ pixels with ImageNet normalization ($\mu = [0.485, 0.456, 0.406]$, $\sigma = [0.229, 0.224, 0.225]$) prepended to all deep architectures. This design maintains end-to-end differentiability for gradient-based attacks while constraining perturbations to the $64 \times 64$ space.

\subsection{Adversarial Example Generation}
\label{sec:adversarial_gen}

\textbf{Surrogate Model.} VGG16~\cite{vgg} with Batch Normalization served as the surrogate, initialized from ImageNet weights and fine-tuned on CIFAR-10 for 10 epochs, achieving 98\% training and 89\% test accuracy.

\textbf{Attack Methods.} Two $\ell_{\infty}$-bounded attacks generated adversarial examples:

\textit{Fast Gradient Sign Method (FGSM)}~\cite{fgsm} computes a single maximal perturbation step:
\begin{equation}
    x_{\text{adv}} = x + \epsilon \cdot \text{sign}(\nabla_x \mathcal{L}(f_\theta(x), y))
\end{equation}

\textit{Projected Gradient Descent (PGD)}~\cite{pgd} iteratively refines perturbations with projection onto the $\ell_{\infty}$ ball:
\begin{equation}
    x^{(t+1)} = \Pi_{B_\epsilon(x)}\left( x^{(t)} + \alpha \cdot \text{sign}(\nabla_{x^{(t)}} \mathcal{L}(f_\theta(x^{(t)}), y)) \right)
\end{equation}
with 10 iterations, step size $\alpha = 2/255$, and random initialization.

\textbf{Perturbation Budgets.} Two budgets were evaluated: $\epsilon = 4/255$ (conservative) and $\epsilon = 8/255$ (aggressive, slightly perceptible). Attacks were applied to the full training set, with adversarial images stored alongside metadata.

\subsection{HOG Feature Extraction}
\label{sec:hog_extraction}

Histogram of Oriented Gradients (HOG)~\cite{hog} features encode local edge orientation distributions via gradient quantization---a fundamentally different representation than learned neural features. All images were converted to grayscale prior to extraction.

Eight configurations (Table~\ref{tab:hog_configs}) systematically varied three parameters: pixels per cell (spatial granularity), orientation bins (angular resolution), and cells per block (normalization scope).

\begin{table}[H]
\centering
\caption{HOG Parameter Configurations}
\label{tab:hog_configs}
\small
\begin{tabular}{cccccc}
\toprule
\textbf{Config} & \textbf{Px/Cell} & \textbf{Orient.} & \textbf{Cells/Blk} & \textbf{$\epsilon$} & \textbf{Dims} \\
\midrule
C1 & 8 & 6 & 1 & 4/255 & 384 \\
C2 & 6 & 6 & 1 & 4/255 & 600 \\
C3 & 10 & 6 & 1 & 4/255 & 216 \\
C4 & 8 & 6 & 2 & 4/255 & 1,176 \\
C5 & 8 & 6 & 3 & 4/255 & 1,944 \\
C6 & 8 & 9 & 1 & 4/255 & 576 \\
C7 & 8 & 3 & 1 & 4/255 & 192 \\
C8 & 8 & 6 & 1 & 8/255 & 384 \\
\bottomrule
\end{tabular}
\end{table}

For each configuration, three parallel datasets were constructed: Original HOG (clean images), FGSM HOG (FGSM-perturbed), and PGD HOG (PGD-perturbed).

\subsection{Hyperparameter Optimization}
\label{sec:hyperparameter}

\textbf{Design Rationale.} Since the primary objective is quantifying adversarial-induced degradation rather than maximizing generalization, we employed single-loop cross-validation. This measures relative accuracy drops without nested CV overhead, with final models trained on full data for stability.

We report accuracy as the sole metric because our objective is measuring relative adversarial degradation rather than optimizing generalization; the absolute accuracy values are secondary to the proportional drops induced by transferred perturbations.

\textbf{Protocol.} Stratified 5-fold CV with grid search was performed on clean HOG features (Configuration C1). Four classical models were tuned:

\begin{itemize}
    \item \textbf{KNN:} $k \in \{3, 5, 9, 15\}$; selected $k=3$
    \item \textbf{Decision Tree:} max\_depth $\in \{3, 5, 10, 15\}$; selected depth$=10$
    \item \textbf{Linear SVM:} PCA $\to$ SGD; PCA variance $\in \{0.70, 0.80, 0.90, 0.95\}$; selected $0.95$
    \item \textbf{Kernel SVM:} PCA(0.80) $\to$ RBF; $C \in \{0.1, 1, 10\}$; selected $C=1$
\end{itemize}

Cross-validation accuracies (0.269 KNN, 0.234 DT, 0.413 LSVM, 0.334 KSVM) reflect HOG's limited representational capacity on CIFAR-10.

\subsection{Neural Network Baselines}
\label{sec:nn_baselines}

\textbf{ANN on HOG Features.} A shallow feedforward network: Input $\to$ 256 $\to$ 64 $\to$ 16 $\to$ 10 (ReLU, softmax). Training: mini-batch SGD, batch size 64, learning rate 0.01, 30 epochs, cross-entropy loss, GPU-accelerated.

\textbf{Deep Transfer Baseline.} AlexNet~\cite{alexnet} was fine-tuned on clean CIFAR-10 (30 epochs, LR 0.001, batch 64), achieving 90\% training and 84\% test accuracy. Evaluating AlexNet on VGG-crafted adversarial images establishes the baseline neural-to-neural transferability for comparison with cross-paradigm transfer to HOG classifiers.

\subsection{Experimental Protocol}
\label{sec:protocol}

Algorithm~\ref{alg:main_experiment} formalizes the evaluation procedure. The core idea is straightforward: for each HOG configuration, we extract three parallel feature sets from clean, FGSM-perturbed, and PGD-perturbed images. Each classifier is trained exclusively on clean HOG features, then evaluated on all three test sets. This design isolates the effect of adversarial perturbations---any accuracy degradation on adversarial sets directly measures transfer vulnerability, since the classifier has never seen perturbed examples during training. Models train on \textit{full} original HOG data since the metric of interest is adversarial degradation, not generalization.

\begin{algorithm}[H]
\caption{Adversarial Robustness Evaluation}
\label{alg:main_experiment}
\small
\begin{algorithmic}[1]
\For{each HOG configuration $C_i$}
    \State Extract: $\mathbf{X}_{\text{orig}}^{(i)}, \mathbf{X}_{\text{fgsm}}^{(i)}, \mathbf{X}_{\text{pgd}}^{(i)}$
    \For{each model $m$ with optimal $\theta^*_m$}
        \State Fit $m$ on $(\mathbf{X}_{\text{orig}}^{(i)}, \mathbf{y})$
        \State Record: Acc$_{\text{orig}}$, Acc$_{\text{fgsm}}$, Acc$_{\text{pgd}}$
    \EndFor
\EndFor
\end{algorithmic}
\end{algorithm}

\subsection{Imperceptibility Verification}

To verify that adversarial perturbations remain imperceptible, we computed cosine similarity between original and adversarial images. Cosine similarity measures the angular distance between two vectors, ranging from $-1$ (opposite) to $+1$ (identical direction). For images represented as flattened pixel vectors, values near 1.0 indicate that the perturbation preserves the overall structure and appearance of the original image:
\begin{equation}
    \text{cos}(x_{\text{orig}}, x_{\text{adv}}) = \frac{\langle x_{\text{orig}}, x_{\text{adv}} \rangle}{\|x_{\text{orig}}\| \cdot \|x_{\text{adv}}\|}
\end{equation}

\begin{figure}[H]
    \centering
    \includegraphics[width=\columnwidth]{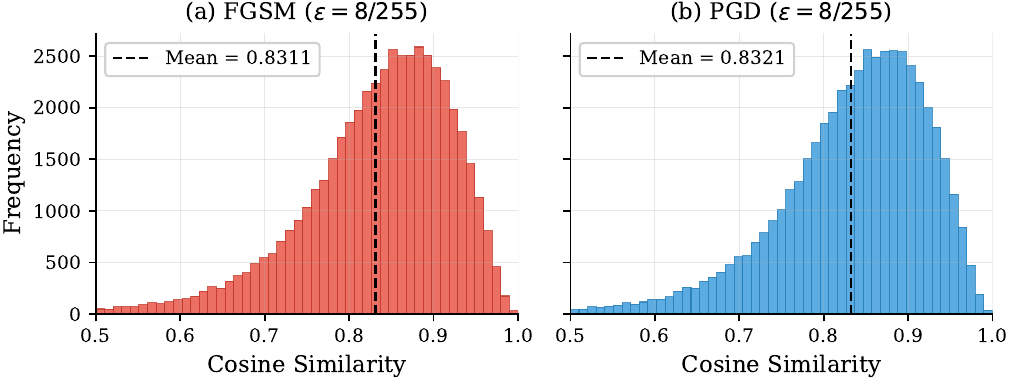}
    \caption{Cosine similarity distributions between original and adversarial images at $\epsilon = 8/255$ (worst-case perturbation budget). Mean similarities of 0.831 (FGSM) and 0.832 (PGD) confirm that adversarial perturbations preserve global image structure even at aggressive attack strengths.}
    \label{fig:similarity_distributions}
\end{figure}

Figure~\ref{fig:similarity_distributions} presents the cosine similarity distributions between original and adversarial images at $\epsilon = 8/255$, representing the upper bound of our perturbation budget. Despite this aggressive setting, both FGSM and PGD maintain mean similarities above 0.83, indicating that the majority of image structure is preserved. The broader distribution compared to smaller $\epsilon$ values reflects the trade-off between attack strength and perceptual fidelity inherent to $L_\infty$-bounded perturbations.

\subsection{Implementation}

Experiments used scikit-learn (classical ML), PyTorch (neural networks, adversarial generation), and scikit-image (HOG). Hardware: NVIDIA RTX 4060, AMD Ryzen 7 8845HS, 16GB RAM. All training performed locally.

Code can be found in the GitHub repository: 
\url{https://github.com/AchrafHsain7/MLHACK_PAPER}

\section{Results}

\subsection{Classical ML Robustness}

Table~\ref{tab:classical_ml_results} presents accuracy across all configurations. All models exhibit substantial degradation, with FGSM causing larger drops than PGD in every case (32/32 configuration-model pairs). Configuration C8 ($\epsilon = 8/255$) produces the most severe degradation.

\begin{table*}[t]
\centering
\caption{Classical ML Accuracy on HOG Features. Bold indicates worst accuracy per model.}
\label{tab:classical_ml_results}
\small
\begin{tabular}{cl|ccc|ccc|ccc|ccc}
\toprule
& & \multicolumn{3}{c|}{\textbf{KNN} ($k$=3)} & \multicolumn{3}{c|}{\textbf{Decision Tree}} & \multicolumn{3}{c|}{\textbf{Linear SVM}} & \multicolumn{3}{c}{\textbf{Kernel SVM}} \\
\textbf{Config} & \textbf{Parameters} & Orig & FGSM & PGD & Orig & FGSM & PGD & Orig & FGSM & PGD & Orig & FGSM & PGD \\
\midrule
C1 & O6-C8-B1-$\epsilon$4 & .543 & .304 & .370 & .352 & .208 & .234 & .420 & .274 & .325 & .743 & .310 & .432 \\
C2 & O6-C6-B1-$\epsilon$4 & .525 & .336 & .386 & .330 & .185 & .211 & .427 & .254 & .315 & .780 & .319 & .454 \\
C3 & O6-C10-B1-$\epsilon$4 & .586 & .324 & .410 & .371 & .231 & .256 & .374 & .282 & .315 & .688 & .333 & .434 \\
C4 & O6-C8-B2-$\epsilon$4 & .647 & .463 & .531 & .394 & .221 & .256 & .491 & .338 & .409 & .848 & .477 & .632 \\
C5 & O6-C8-B3-$\epsilon$4 & .716 & .597 & .652 & .412 & .202 & .247 & .497 & .361 & .428 & .862 & .598 & .730 \\
C6 & O9-C8-B1-$\epsilon$4 & .552 & .329 & .390 & .356 & .212 & .235 & .451 & .284 & .343 & .793 & .368 & .483 \\
C7 & O3-C8-B1-$\epsilon$4 & .558 & .247 & .347 & .347 & .208 & .235 & .302 & .228 & .250 & .601 & .284 & .370 \\
C8 & O6-C8-B1-$\epsilon$8 & .543 & \textbf{.139} & .259 & .352 & \textbf{.149} & .200 & .412 & \textbf{.177} & .272 & .743 & \textbf{.132} & .258 \\
\bottomrule
\end{tabular}
\end{table*}

\begin{figure}[H]
    \centering
    \includegraphics[width=\columnwidth]{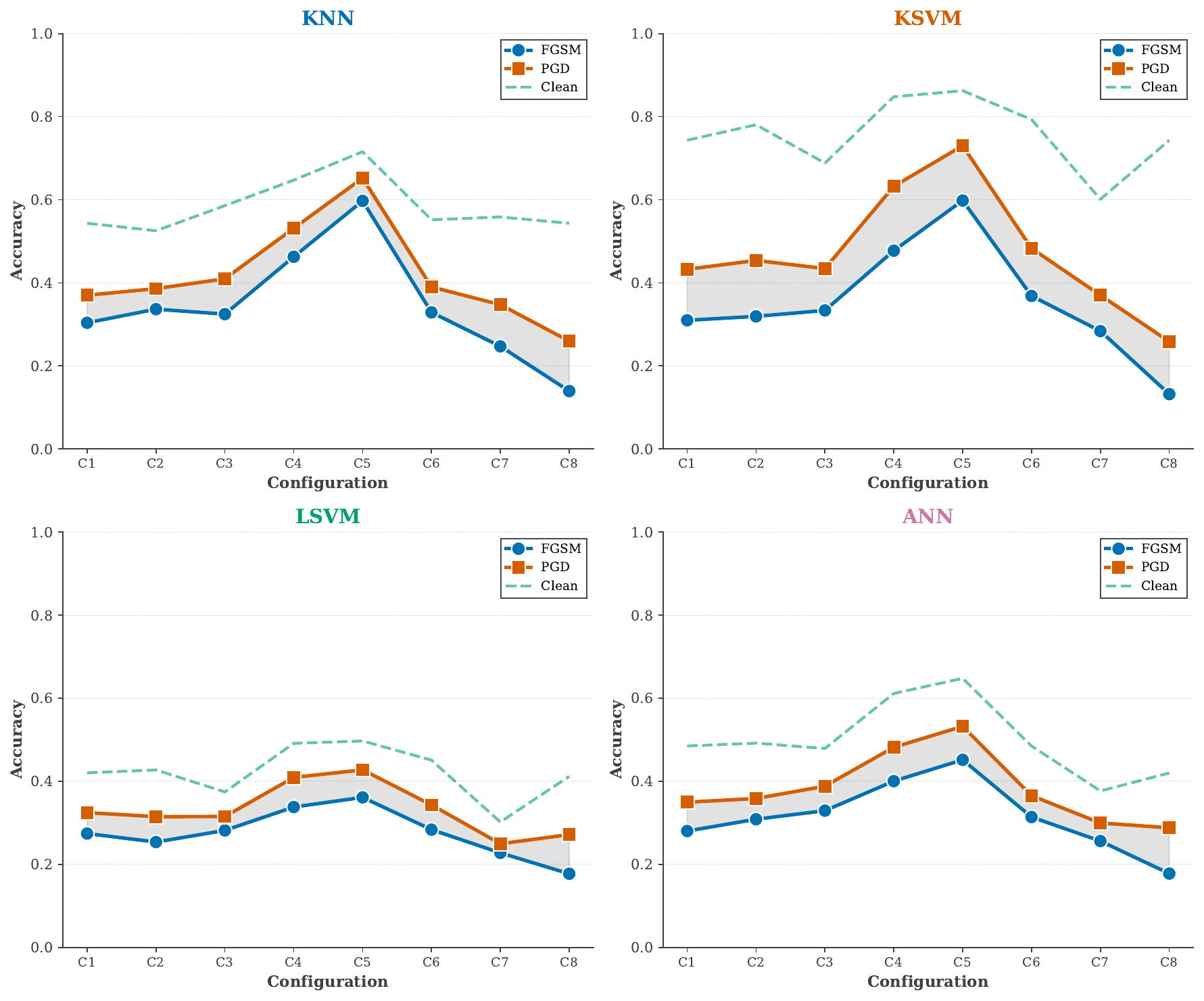}
    \caption{Sensitivity of HOG-based classifiers to adversarial perturbation magnitude across configurations. Shaded regions highlight the FGSM-PGD gap.}
    \label{fig:config_sensitivity}
\end{figure}

Figure~\ref{fig:config_sensitivity} illustrates configuration-wise accuracy trajectories for K-NN, RBF-SVM, Linear SVM, and ANN under FGSM, PGD, and clean conditions. The shaded region between FGSM and PGD curves highlights the consistent gap where PGD maintains higher accuracy than FGSM across configurations---visual evidence of the attack hierarchy reversal. All classifiers exhibit parallel response patterns: peak robustness at C5 (Block=3), moderate performance at C4 (Block=2), and severe degradation at C8 ($\epsilon=8$). This consistency suggests that HOG parameter sensitivity generalizes across classifier architectures.

\subsection{ANN Robustness on HOG Features}

Table~\ref{tab:ann_results} reports ANN performance. Vulnerability patterns mirror classical models, with FGSM consistently causing greater degradation than PGD across configurations.

\begin{table}[H]
\centering
\caption{ANN Accuracy on HOG Features}
\label{tab:ann_results}
\small
\begin{tabular}{clccc}
\toprule
\textbf{Config} & \textbf{Parameters} & \textbf{Orig} & \textbf{FGSM} & \textbf{PGD} \\
\midrule
C1 & O6-C8-B1-$\epsilon$4 & 0.485 & 0.280 & 0.349 \\
C2 & O6-C6-B1-$\epsilon$4 & 0.492 & 0.309 & 0.359 \\
C3 & O6-C10-B1-$\epsilon$4 & 0.479 & 0.329 & 0.388 \\
C4 & O6-C8-B2-$\epsilon$4 & 0.612 & 0.400 & 0.482 \\
C5 & O6-C8-B3-$\epsilon$4 & 0.648 & 0.452 & 0.532 \\
C6 & O9-C8-B1-$\epsilon$4 & 0.485 & 0.314 & 0.365 \\
C7 & O3-C8-B1-$\epsilon$4 & 0.377 & 0.256 & 0.300 \\
C8 & O6-C8-B1-$\epsilon$8 & 0.420 & \textbf{0.178} & 0.288 \\
\bottomrule
\end{tabular}
\end{table}

\subsection{Deep Learning Transfer Baseline}

Table~\ref{tab:cnn_transfer} compares on-surrogate attack success with cross-architecture transfer. VGG suffers near-complete collapse under its own attacks ($\sim$80\% drop for FGSM, $\sim$99\% for PGD). AlexNet shows moderate accuracy drop ($\sim$13\% FGSM, $\sim$7\% PGD), confirming partial but meaningful transfer. Notably, PGD---the stronger on-surrogate attack---transfers \textit{less} effectively than FGSM, supporting the hypothesis that iterative attacks overfit to surrogate-specific boundaries.

\begin{table}[H]
\centering
\caption{CNN Transferability: VGG (surrogate) $\to$ AlexNet (target)}
\label{tab:cnn_transfer}
\small
\begin{tabular}{lccc}
\toprule
\textbf{Model} & \textbf{Clean} & \textbf{FGSM} & \textbf{PGD} \\
\midrule
VGG (surrogate) & 0.98 & $\sim$0.18 & $\sim$0.01 \\
AlexNet (target) & 0.90 & $\sim$0.77 & $\sim$0.83 \\
\bottomrule
\end{tabular}
\end{table}

\subsection{Cross-Paradigm Comparison}

Table~\ref{tab:cross_paradigm} synthesizes results using configuration C5 (best-performing HOG setup). Relative drop is computed as $(\text{Orig} - \text{FGSM}) / \text{Orig} \times 100\%$. HOG-based models suffer comparable or greater relative degradation than the neural transfer baseline, demonstrating that feature engineering provides no inherent protection against transferred adversarial perturbations.

\begin{table}[H]
\centering
\caption{Cross-Paradigm Adversarial Degradation (Config C5)}
\label{tab:cross_paradigm}
\small
\begin{tabular}{llccr}
\toprule
\textbf{Pipeline} & \textbf{Model} & \textbf{Orig} & \textbf{FGSM} & \textbf{Rel. Drop} \\
\midrule
\multirow{2}{*}{HOG+Classical} 
 & KSVM & .862 & .598 & 30.7\% \\
 & KNN & .716 & .597 & 16.6\% \\
\midrule
HOG+Neural & ANN & .648 & .452 & 30.3\% \\
\midrule
End-to-End & AlexNet & .90 & .77 & 14.4\% \\
\bottomrule
\end{tabular}
\end{table}

\begin{figure*}[t]
    \centering
    \includegraphics[width=\textwidth]{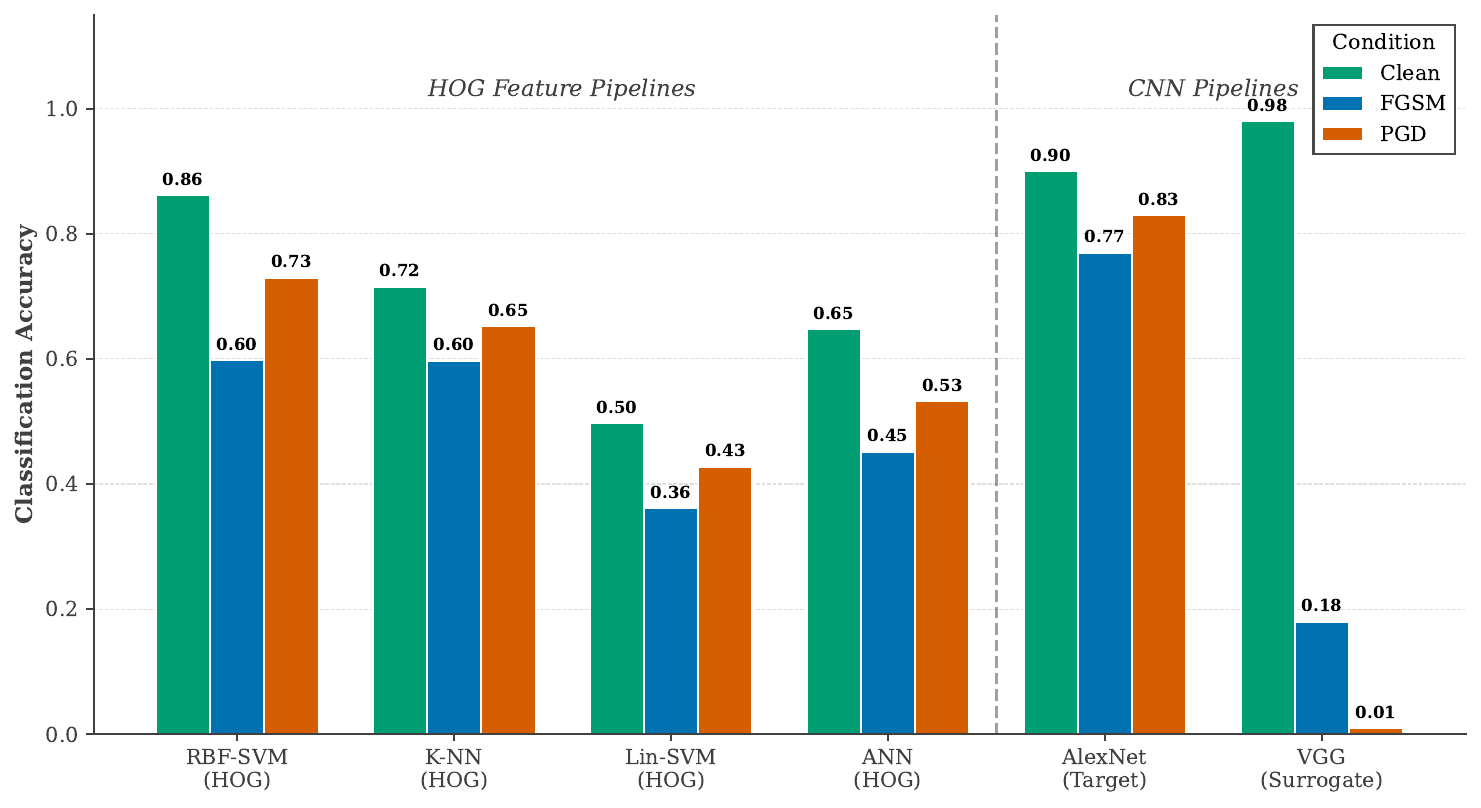}
    \caption{Comparative analysis of adversarial robustness across feature-engineered and end-to-end classification pipelines. Classification accuracy of HOG-based pipelines (RBF-SVM, K-NN, Linear SVM, and ANN) compared against CNN transfer baselines (AlexNet as target, VGG as surrogate) under clean, FGSM, and PGD conditions. All HOG classifiers use Configuration C5 (block size = 3).}
    \label{fig:cross_paradigm_bar}
\end{figure*}

Figure~\ref{fig:cross_paradigm_bar} provides a visual comparison of classification accuracy across HOG-based pipelines (RBF-SVM, K-NN, Linear SVM, and ANN) compared against the CNN transfer baseline (AlexNet as target, VGG as surrogate) under clean, FGSM, and PGD conditions. All HOG classifiers use Configuration C5 (block size = 3), which achieved optimal performance. The results demonstrate that adversarial perturbations crafted on the VGG surrogate transfer effectively to HOG-based classifiers, causing 17--31\% relative accuracy degradation---comparable to the 14\% degradation observed in the AlexNet neural transfer baseline. This finding refutes the hypothesis that gradient-based feature engineering provides inherent robustness against CNN-crafted adversarial examples.

\begin{figure}[h]
    \centering
    \includegraphics[width=\columnwidth]{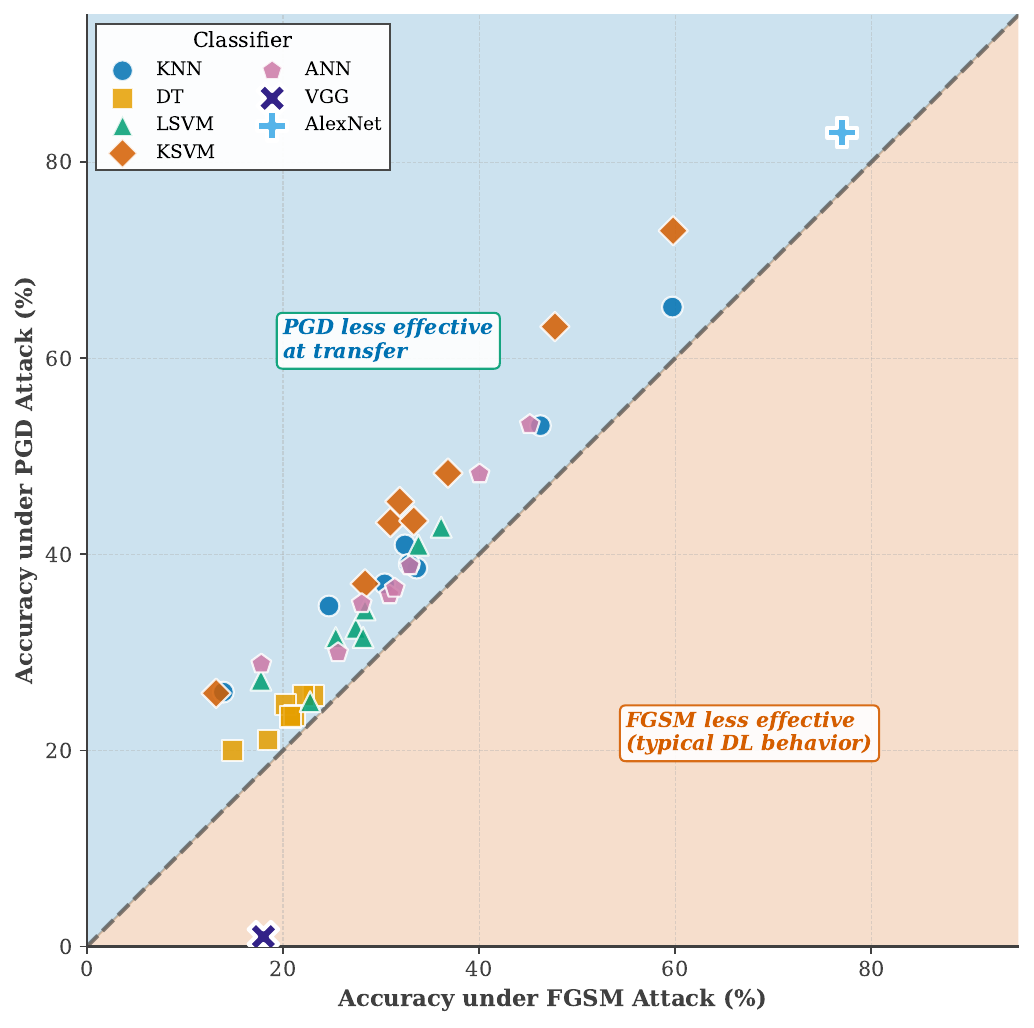}
    \caption{FGSM vs.\ PGD transfer effectiveness to HOG feature space: evidence of attack hierarchy reversal. All points lie above the diagonal, indicating FGSM causes greater accuracy degradation than PGD across all configurations.}
    \label{fig:fgsm_vs_pgd_scatter}
\end{figure}

Figure~\ref{fig:fgsm_vs_pgd_scatter} presents a scatter plot comparing classifier accuracy under FGSM versus PGD attacks across all experimental configurations. Points above the diagonal indicate higher accuracy under PGD than FGSM, meaning PGD transfers less effectively. Remarkably, 100\% of HOG-based classifier-configuration pairs ($n=40$) fall above the diagonal, demonstrating a consistent reversal of the typical deep learning attack hierarchy where PGD dominates FGSM. We hypothesize that PGD's iterative optimization exploits surrogate-specific high-level features that do not survive HOG's gradient quantization, while FGSM's single-step perturbation more directly corrupts the low-level edge structures preserved by HOG.

\section{Discussion}

\subsection{Hypothesis Evaluation}

The hypothesis that HOG's gradient quantization might discard adversarial signals is \textbf{strongly refuted}. All classifiers exhibited substantial accuracy degradation under transferred attacks, with relative drops ranging from 16.6\% to 59.1\% at $\epsilon = 4/255$ depending on configuration and classifier type.

\subsection{Perturbation Budget Sensitivity}

Comparing C1 ($\epsilon=4/255$) and C8 ($\epsilon=8/255$), which share identical HOG parameters, reveals severe nonlinear sensitivity. Table~\ref{tab:eps_sensitivity} reports the accuracy and relative drop (computed as percentage decrease from original). Doubling the perturbation budget increases relative degradation by up to 30 percentage points.

\begin{table}[H]
\centering
\caption{FGSM Accuracy by Perturbation Budget (C1 vs C8)}
\label{tab:eps_sensitivity}
\small
\begin{tabular}{lcccc}
\toprule
& \multicolumn{2}{c}{\textbf{$\epsilon$=4/255 (C1)}} & \multicolumn{2}{c}{\textbf{$\epsilon$=8/255 (C8)}} \\
\textbf{Model} & FGSM & Rel.Drop & FGSM & Rel.Drop \\
\midrule
KSVM & .310 & 58.3\% & .132 & 82.3\% \\
KNN & .304 & 44.1\% & .139 & 74.4\% \\
LSVM & .274 & 34.8\% & .177 & 56.9\% \\
DT & .208 & 41.0\% & .149 & 57.7\% \\
\bottomrule
\end{tabular}
\end{table}

\begin{figure}[H]
    \centering
    \includegraphics[width=\columnwidth]{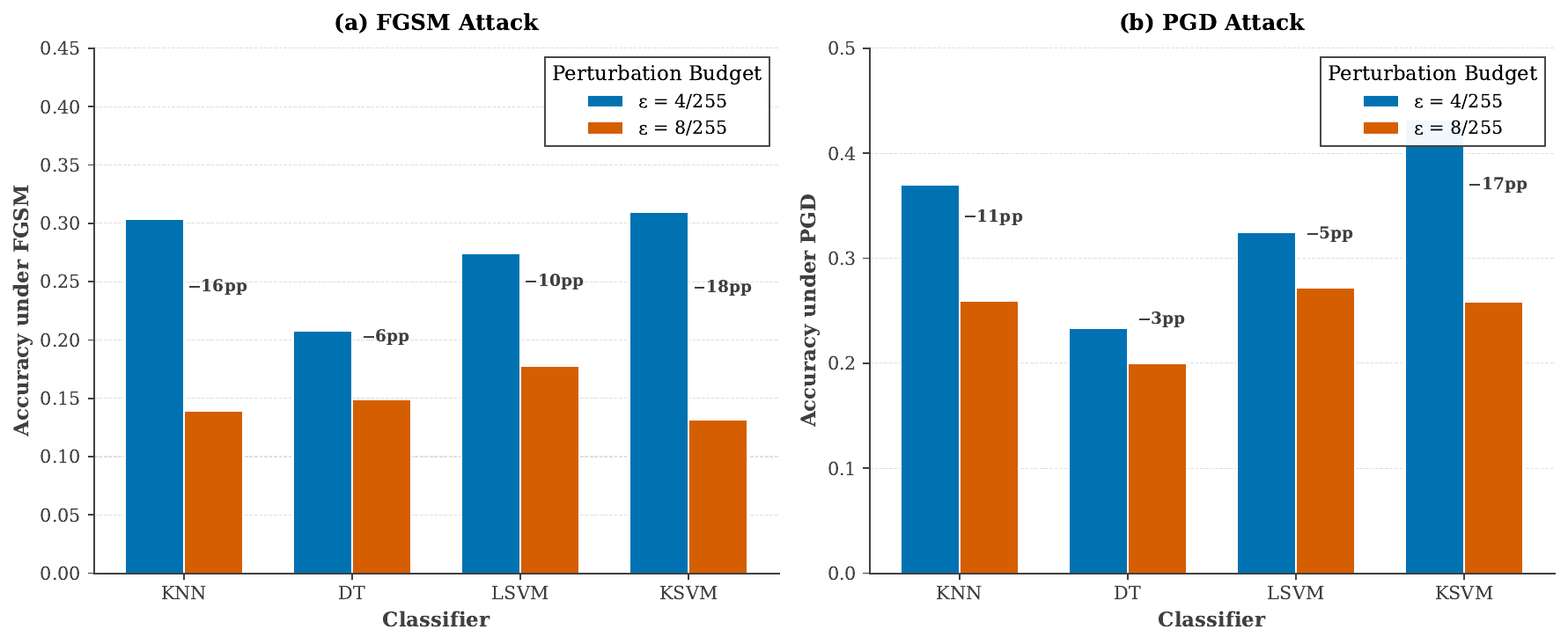}
    \caption{Sensitivity of HOG-based classifiers to adversarial perturbation magnitude: $\epsilon = 4/255$ vs.\ $\epsilon = 8/255$.}
    \label{fig:eps_sensitivity}
\end{figure}

Figure~\ref{fig:eps_sensitivity} illustrates the comparison of classifier accuracy under FGSM and PGD attacks at the two perturbation budgets. Doubling the perturbation budget produces disproportionately larger accuracy degradation, with drops of 11--18 percentage points for FGSM and 7--17 percentage points for PGD. RBF-SVM exhibits the most severe sensitivity, dropping from 0.31 to 0.13 ($-58\%$) under FGSM at $\epsilon = 8/255$. This nonlinear relationship suggests that HOG features become increasingly corrupted as perturbation magnitude increases, with no evident plateau or saturation effect within the tested range.

\subsection{FGSM vs.\ PGD Transfer}

Across all 32 classical ML configuration-model combinations, FGSM caused larger accuracy drops than PGD (100\% of cases). This contradicts the typical deep learning hierarchy where PGD dominates. We hypothesize that PGD's iterative optimization exploits surrogate-specific high-level features that don't survive HOG's gradient-based encoding, while FGSM's single-step perturbation more directly corrupts the low-level edge structures HOG preserves.

\subsection{HOG Parameter Impact}

Increasing cells-per-block (B) consistently improved both clean and robust accuracy. Table~\ref{tab:block_impact} shows FGSM accuracy across block sizes (C1, C4, C5 share O=6, C=8, $\epsilon$=4). Larger blocks induce spatial averaging during contrast normalization, potentially smoothing high-frequency perturbations. However, even B=3 suffered 17--31\% relative accuracy drops.

\begin{table}[H]
\centering
\caption{Effect of Block Size on FGSM Robustness}
\label{tab:block_impact}
\small
\begin{tabular}{lcccccc}
\toprule
& \multicolumn{2}{c}{\textbf{B=1 (C1)}} & \multicolumn{2}{c}{\textbf{B=2 (C4)}} & \multicolumn{2}{c}{\textbf{B=3 (C5)}} \\
\textbf{Model} & Orig & FGSM & Orig & FGSM & Orig & FGSM \\
\midrule
KSVM & .743 & .310 & .848 & .477 & .862 & .598 \\
KNN & .543 & .304 & .647 & .463 & .716 & .597 \\
\bottomrule
\end{tabular}
\end{table}

\begin{figure}[H]
    \centering
    \includegraphics[width=\columnwidth]{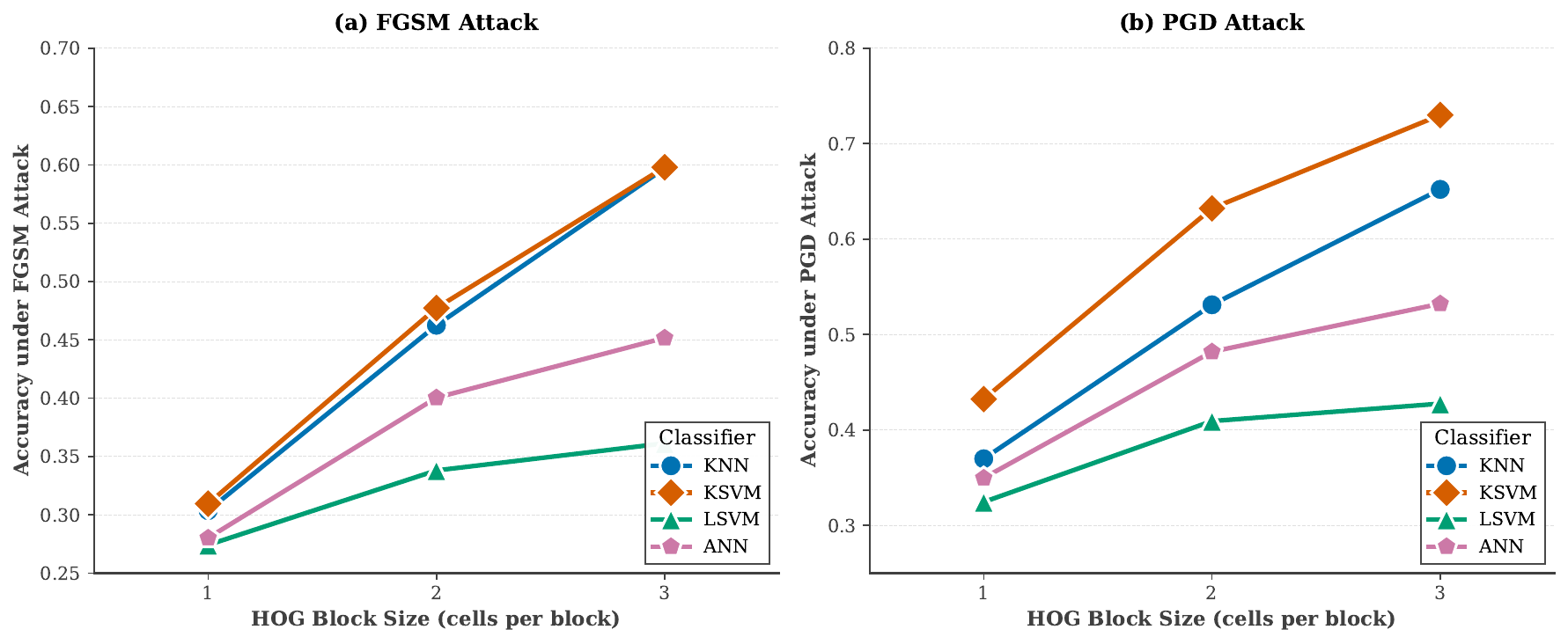}
    \caption{Effect of HOG block size on adversarial robustness under FGSM and PGD attacks.}
    \label{fig:block_size}
\end{figure}

Figure~\ref{fig:block_size} visualizes adversarial accuracy as a function of HOG block size for K-NN, RBF-SVM, and Linear SVM classifiers under FGSM and PGD attacks. Increasing block size from $1\times1$ to $3\times3$ yields substantial robustness improvements: K-NN improves from 0.30 to 0.60 (+97\%) under FGSM, while RBF-SVM improves from 0.31 to 0.60 (+93\%). Larger blocks induce spatial averaging during L2 contrast normalization, which attenuates high-frequency adversarial perturbations. However, even the optimal configuration (B=3) suffers 17--31\% relative accuracy degradation, indicating that block normalization provides partial mitigation but cannot fully prevent adversarial transfer.

\subsection{Multi-Metric Robustness Analysis}

\begin{figure}[H]
    \centering
    \includegraphics[width=\columnwidth]{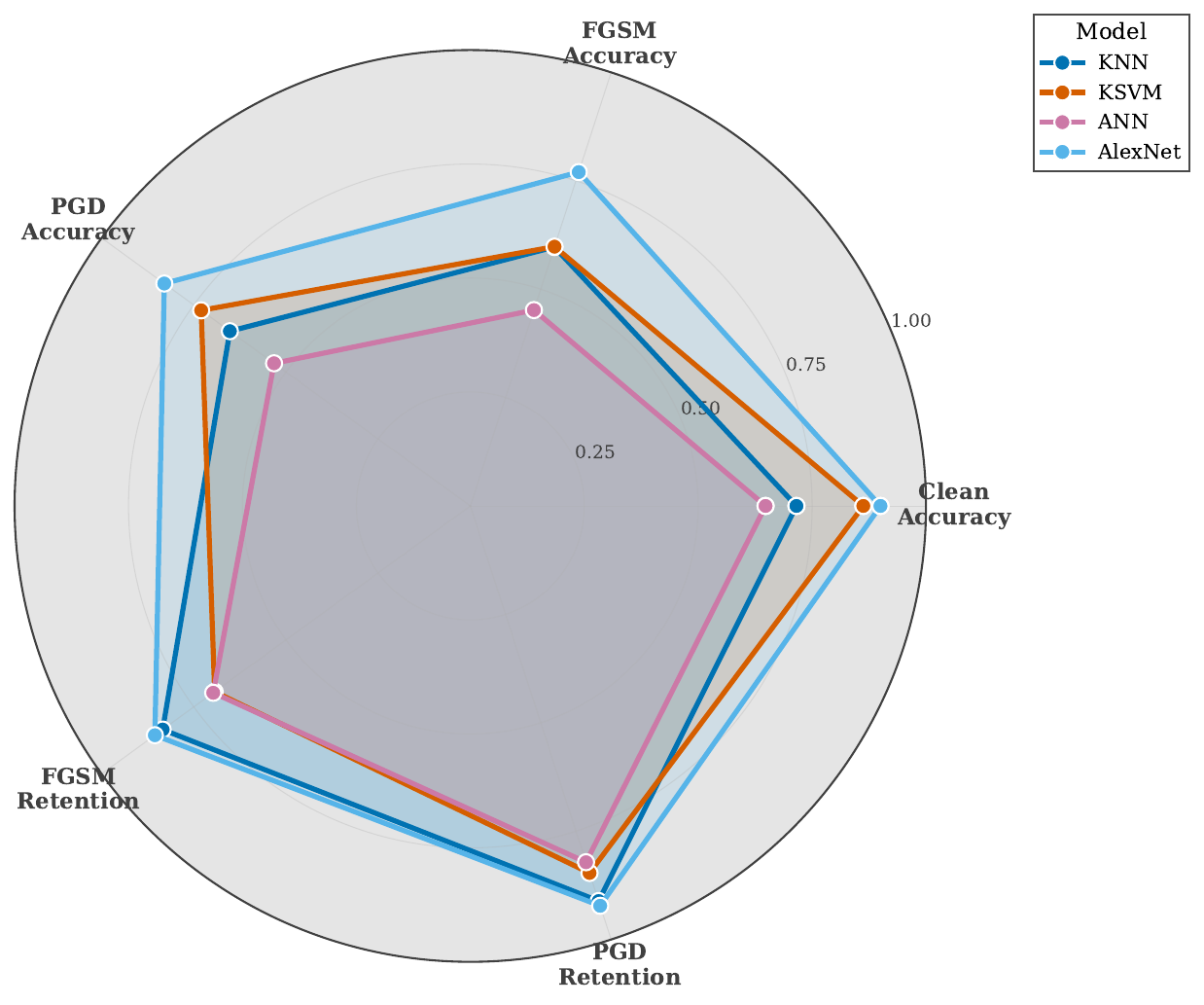}
    \caption{Radar chart comparison of classifier robustness profiles across five performance metrics.}
    \label{fig:radar}
\end{figure}

Figure~\ref{fig:radar} provides a multi-dimensional comparison of K-NN, RBF-SVM, ANN (all HOG-based, Configuration C5), and AlexNet (CNN transfer baseline) across five metrics: clean accuracy, FGSM accuracy, PGD accuracy, FGSM retention, and PGD retention. AlexNet achieves the largest radar area, indicating superior overall robustness, with notably high retention rates (86--92\%). Among HOG classifiers, K-NN and RBF-SVM show similar profiles with high PGD retention but lower FGSM retention---consistent with the attack hierarchy reversal. ANN exhibits the smallest area, suggesting shallow neural networks on HOG features inherit vulnerability without gaining the retention advantages of distance-based classifiers.

\begin{figure}[H]
    \centering
    \includegraphics[width=\columnwidth]{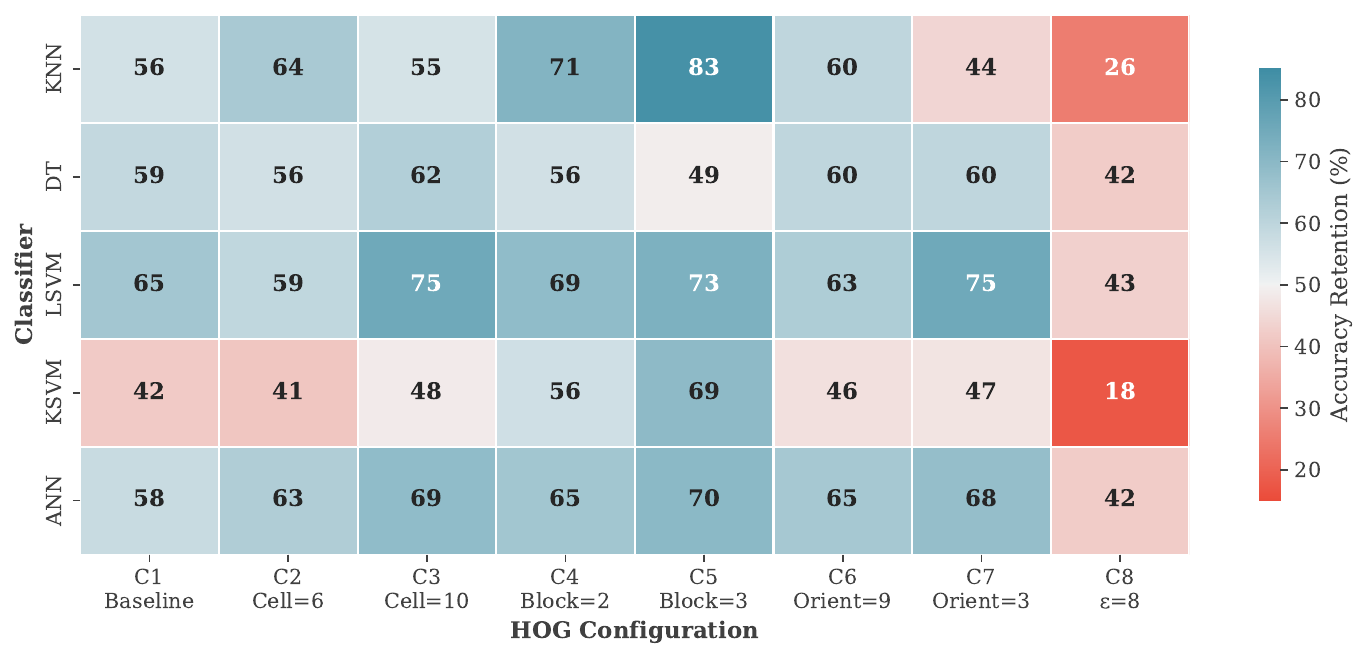}
    \caption{FGSM accuracy retention heatmap across HOG configurations and classifier types.}
    \label{fig:retention_heatmap}
\end{figure}

Figure~\ref{fig:retention_heatmap} presents a heatmap showing the percentage of original accuracy retained under FGSM attack for each classifier-configuration pair. Values represent $(\text{FGSM accuracy} / \text{Clean accuracy}) \times 100\%$. Configuration C5 (Block=3) achieves the highest retention across all classifiers (69--83\%), while C8 ($\epsilon=8$) produces the lowest retention (18--43\%). Decision Tree exhibits consistently moderate retention ($\sim$40--60\%) regardless of configuration, suggesting its vulnerability stems from architectural rather than feature-level factors.

\subsection{Correlation Analysis}

\begin{figure}[H]
    \centering
    \includegraphics[width=\columnwidth]{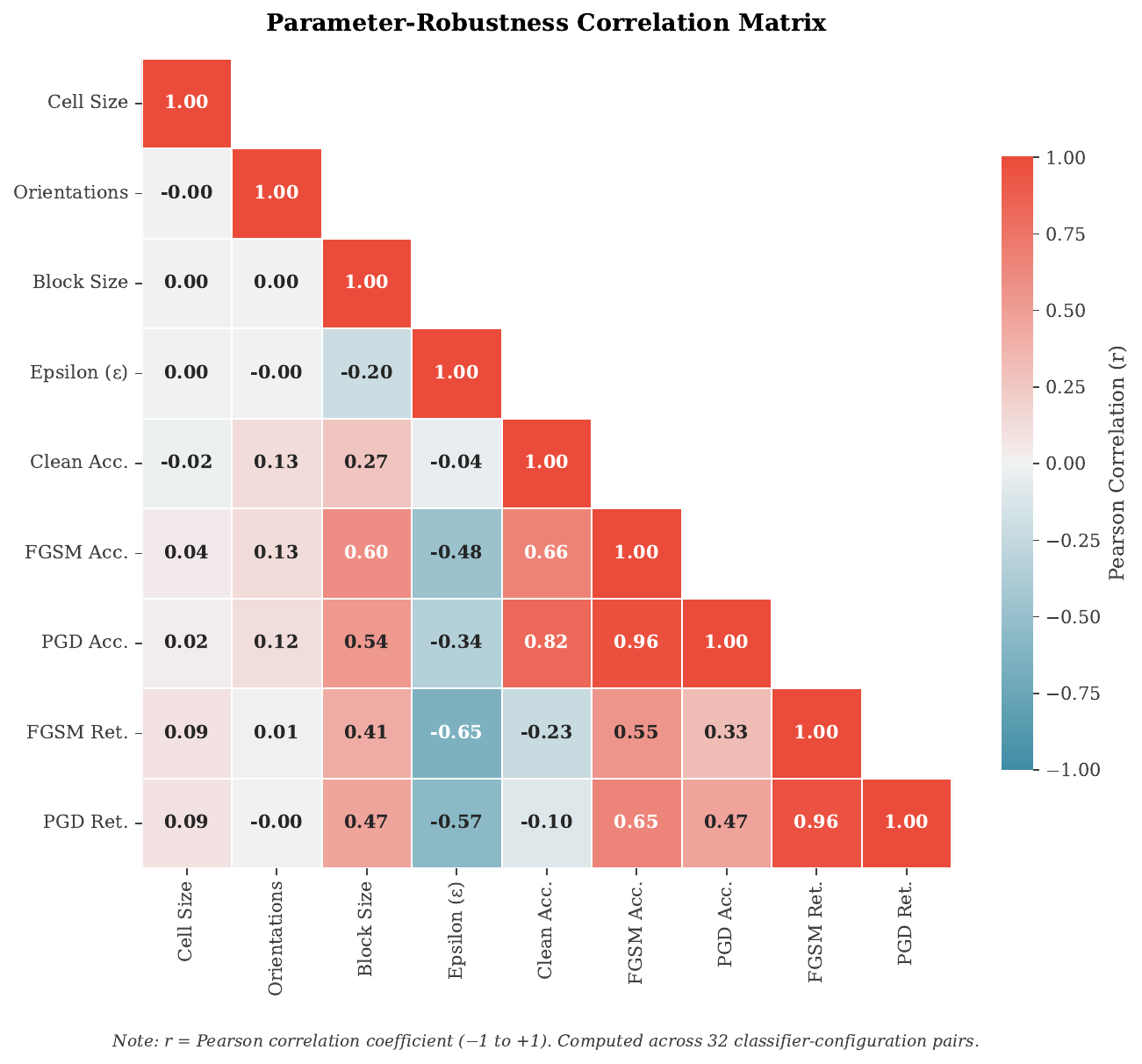}
    \caption{Pearson correlation matrix of HOG parameters and adversarial robustness metrics.}
    \label{fig:correlation}
\end{figure}

Figure~\ref{fig:correlation} presents a correlation matrix showing relationships between HOG configuration parameters (cell size, orientations, block size, epsilon) and robustness metrics (clean accuracy, FGSM/PGD accuracy, FGSM/PGD retention). Computed across 32 classifier-configuration pairs, key findings emerge: (1) Block size shows strong positive correlation with FGSM accuracy ($r = 0.60$) and retention ($r = 0.41$); (2) Epsilon shows strong negative correlation with FGSM retention ($r = -0.65$); (3) Cell size and orientations show negligible correlation with robustness ($|r| < 0.15$). Note that correlations measure configuration sensitivity, not measurement uncertainty from repeated trials.

\subsection{Comparison with Prior HOG Adversarial Studies}

Our findings both complement and extend prior work examining HOG features in adversarial contexts.

\textbf{Contrast with Lateralized Defenses.} Siddique et al.~\cite{siddique2020} achieved 81\% accuracy against FGSM attacks through a novel lateralized architecture combining HOG and SIFT features, compared to 70\% for standard VGG. Our results suggest that this improvement likely stems from their heterogeneous feature combination and lateralized processing rather than HOG's intrinsic properties---when HOG features are used alone without architectural innovations, transferred attacks cause 17--31\% relative degradation even under optimal configurations.

\textbf{Consistency with Direct Attack Vulnerability.} Mohammed et al.~\cite{mohammed2025} found that HOG-based classifiers exhibit substantial vulnerability when attacks are crafted directly on the HOG pipeline. Our results extend this finding to the transfer setting: even attacks optimized against CNN loss landscapes---without any knowledge of the HOG feature extraction---successfully degrade HOG classifier performance. This indicates that adversarial perturbations exploit fundamental image properties (edge structures) rather than model-specific decision boundaries.

\textbf{Isolating HOG's Contribution.} Lal et al.~\cite{lal2021} achieved 99.9\% accuracy against FGSM and DeepFool attacks using a combination of HOG, LBP, SFTA, and deep features with adversarial training. Our results provide critical context for interpreting this finding: HOG features alone, without adversarial training or multi-feature fusion, suffer 17--59\% relative degradation under transferred attacks. This suggests that the robustness observed in Lal's work derives primarily from adversarial training and the complementary information in deep features---not from any inherent protective property of HOG's gradient quantization. Our work thus isolates and refutes the hypothesis that HOG features provide natural adversarial robustness.

\subsection{Threats to Validity and Future Work}

Several limitations constrain the generalizability of our findings, which we acknowledge as directions for future investigation.

\textbf{External Validity.} Our experiments used a single dataset (CIFAR-10) and a single surrogate architecture (VGG16). While CIFAR-10 is a standard benchmark, future work should validate these findings on higher-resolution datasets (e.g., ImageNet) and diverse surrogate architectures (e.g., ResNet, Vision Transformers) to confirm that the observed transferability patterns generalize across domains and model families.

\textbf{Statistical Rigor.} Due to hardware and time constraints, we did not perform repeated trials with different random seeds or report confidence intervals. Future work should incorporate statistical significance testing to quantify the reliability of observed accuracy differences across configurations.

\textbf{Surrogate Diversity.} We employed only VGG16 as the surrogate model. Investigating whether attacks crafted on architecturally diverse surrogates (e.g., ResNet, EfficientNet, ViT) exhibit different transfer characteristics to HOG-based classifiers remains an open question.

\textbf{Metric Selection.} We focused exclusively on accuracy as our evaluation metric. This choice is deliberate: our objective is quantifying adversarial-induced degradation rather than optimizing generalization performance. Per-class metrics (precision, recall, F1) and confusion matrix analysis could reveal whether certain CIFAR-10 classes are disproportionately affected by transferred perturbations---a direction we leave for future investigation.

\textbf{Defense Mechanisms.} This study characterizes vulnerability without exploring mitigations. Future work could investigate whether input preprocessing (e.g., JPEG compression, spatial smoothing) or adversarial training on HOG features can reduce transfer susceptibility.

\section{HOG Configuration Analysis for CIFAR-10}

Beyond adversarial robustness, our experiments provide guidance on HOG parameterization for CIFAR-10. Table~\ref{tab:hog_clean} summarizes clean accuracy across all configurations, organized by the parameter being varied. The table reveals three key trends: (1) block size exhibits the strongest influence, with B=3 achieving the highest accuracy across all classifiers; (2) cell size shows a non-monotonic relationship, with C=10 favoring KNN and DT but hurting SVM variants; and (3) orientation bins primarily affect SVM classifiers, with O=9 yielding modest improvements for LSVM and KSVM while leaving KNN and DT largely unchanged. These findings are consistent with recommendations from the original HOG paper~\cite{hog}, which emphasized block normalization as the most critical parameter for capturing robust gradient statistics.

\begin{table}[H]
\centering
\caption{Clean Accuracy by HOG Configuration}
\label{tab:hog_clean}
\small
\begin{tabular}{clcccc}
\toprule
\textbf{Config} & \textbf{Varied Param} & \textbf{KNN} & \textbf{DT} & \textbf{LSVM} & \textbf{KSVM} \\
\midrule
C1 & B=1 (baseline) & .543 & .352 & .420 & .743 \\
C4 & B=2 & .647 & .394 & .491 & .848 \\
C5 & B=3 & \textbf{.716} & \textbf{.412} & \textbf{.497} & \textbf{.862} \\
\midrule
C2 & C=6 & .525 & .330 & .427 & .780 \\
C1 & C=8 & .543 & .352 & .420 & .743 \\
C3 & C=10 & .586 & .371 & .374 & .688 \\
\midrule
C7 & O=3 & .558 & .347 & .302 & .601 \\
C1 & O=6 & .543 & .352 & .420 & .743 \\
C6 & O=9 & .552 & .356 & .451 & .793 \\
\bottomrule
\end{tabular}
\end{table}

\section{Conclusion}

This study addressed a critical gap in adversarial machine learning: whether classical, feature-engineered pipelines---specifically HOG-based classifiers---offer inherent robustness against adversarial examples crafted on deep neural network surrogates. The hypothesis that HOG's coarse gradient quantization might discard high-frequency adversarial signals was \textbf{strongly refuted}. Our systematic evaluation across eight HOG configurations, five classifier architectures, and two attack methods yields the following principal findings:

\begin{enumerate}
    \item \textbf{HOG provides no inherent defense against transferred attacks.} All classifiers suffered substantial accuracy degradation under FGSM at $\epsilon = 4/255$, ranging from 16.6\% to 59.1\% relative drop depending on configuration and classifier. Even under optimal HOG parameters (Configuration C5), degradation ranged from 16.6\% to 30.7\% for the best-performing classifiers---comparable to or exceeding the 14.4\% observed in neural-to-neural transfer (VGG$\to$AlexNet). This demonstrates that adversarial vulnerability is not an artifact of end-to-end differentiability but a fundamental property of image classification systems.
    
    \item \textbf{Attack hierarchy reversal in cross-paradigm transfer.} Contrary to the established deep learning pattern where iterative PGD dominates single-step FGSM, we observed FGSM causing larger accuracy drops than PGD in 100\% of classical ML configuration-model pairs (32/32). This suggests PGD overfits to surrogate-specific high-level features that do not survive HOG's gradient-based encoding.
    
    \item \textbf{Vulnerability scales nonlinearly with perturbation budget.} Doubling $\epsilon$ from $4/255$ to $8/255$ increased relative accuracy degradation by up to 30 percentage points, with no saturation observed within the tested range.
    
    \item \textbf{Block normalization provides partial but insufficient mitigation.} Increasing cells-per-block from 1 to 3 improved FGSM retention from 42\% to 69\% for KSVM, yet substantial vulnerability persists even under optimal configuration.
\end{enumerate}

These findings carry significant implications for security-critical deployments: the assumption that classical ML pipelines operating on hand-crafted features are immune to neural-network-derived adversarial attacks is unfounded. Adversarial transferability transcends computational paradigms, necessitating explicit robustness considerations regardless of the classification architecture employed.

\bibliographystyle{IEEEtran}
\bibliography{ref}

@inproceedings{fgsm,
  author    = {Goodfellow, Ian J. and Shlens, Jonathon and Szegedy, Christian},
  title     = {Explaining and Harnessing Adversarial Examples},
  booktitle = {International Conference on Learning Representations (ICLR)},
  year      = {2015}
}

@inproceedings{pgd,
  author    = {Madry, Aleksander and Makelov, Aleksandar and Schmidt, Ludwig and Tsipras, Dimitris and Vladu, Adrian},
  title     = {Towards Deep Learning Models Resistant to Adversarial Attacks},
  booktitle = {International Conference on Learning Representations (ICLR)},
  year      = {2018}
}

@inproceedings{yang2020,
  author    = {Yang, Chawin and Kortylewski, Adam and Xie, Cihang and Cao, Yue and Yuille, Alan},
  title     = {PatchAttack: A Black-box Texture-based Attack with Reinforcement Learning},
  booktitle = {Proceedings of the European Conference on Computer Vision (ECCV)},
  pages     = {681--698},
  year      = {2020}
}

@article{agarwal2022,
  author  = {Agarwal, Anshuman and Ratha, Nalini and Vatsa, Mayank and Singh, Richa},
  title   = {Crafting Adversarial Perturbations via Transformed Image Component Swapping},
  journal = {IEEE Transactions on Image Processing},
  volume  = {31},
  pages   = {7338--7349},
  year    = {2022}
}

@inproceedings{mahmood2021,
  author    = {Mahmood, Khalid and Mahmood, Rafay and Van Dijk, Marten},
  title     = {On the Robustness of Vision Transformers to Adversarial Examples},
  booktitle = {Proceedings of the IEEE/CVF International Conference on Computer Vision (ICCV)},
  pages     = {7838--7847},
  year      = {2021}
}

@inproceedings{maria2019,
  author    = {James, Maria and Mruthula, M. and Bhaskaran, V. and Asha, S.},
  title     = {Evasion Attacks on SVM Classifier},
  booktitle = {Proceedings of the 9th International Conference on Advances in Computing and Communication (ICACC)},
  pages     = {125--129},
  year      = {2019}
}

@inproceedings{waseda2023,
  author    = {Waseda, Fuki and Nishikawa, Shusaku and Le, Thanh Nguyen and Nguyen, Huy H. and Echizen, Isao},
  title     = {A Closer Look at the Transferability of Adversarial Examples: How They Fool Different Models Differently},
  booktitle = {Proceedings of the IEEE/CVF Winter Conference on Applications of Computer Vision (WACV)},
  pages     = {1360--1368},
  year      = {2023}
}

@misc{hayes2017,
  author = {Hayes, Jamie and Danezis, George},
  title  = {Machine Learning as an Adversarial Service: Learning Black-Box Adversarial Examples},
  note   = {arXiv preprint arXiv:1708.05207},
  year   = {2017}
}

@article{weng2025,
  author  = {Weng, Jun and Luo, Zhipeng and Li, Shaofeng},
  title   = {Improving Transferable Targeted Adversarial Attack via Normalized Logit Calibration and Truncated Feature Mixing},
  journal = {IEEE Transactions on Information Forensics and Security},
  year    = {2025}
}

@inproceedings{guo2020,
  author    = {Guo, Yuyang and Li, Qizhang and Chen, Hao},
  title     = {Backpropagating Linearly Improves Transferability of Adversarial Examples},
  booktitle = {Advances in Neural Information Processing Systems (NeurIPS)},
  volume    = {33},
  pages     = {85--95},
  year      = {2020}
}

@article{gao2021,
  author  = {Gao, Lianli and Huang, Zhen and Song, Jingkuan and Yang, Yi and Shen, Heng Tao},
  title   = {Push \& Pull: Transferable Adversarial Examples with Attentive Attack},
  journal = {IEEE Transactions on Multimedia},
  volume  = {24},
  pages   = {2329--2338},
  year    = {2021}
}

@article{he2022,
  author  = {He, Zhuang and Duan, Yuntao and Zhang, Weiming and Zou, Jun and Wang, Yushu and Pan, Zhen},
  title   = {Boosting Adversarial Attacks with Transformed Gradient},
  journal = {Computers \& Security},
  volume  = {118},
  pages   = {102720},
  year    = {2022}
}

@techreport{cifar10,
  author = {Krizhevsky, Alex},
  title = {Learning Multiple Layers of Features from Tiny Images},
  institution = {University of Toronto},
  year = {2009}
}

@inproceedings{vgg,
  author    = {Simonyan, Karen and Zisserman, Andrew},
  title     = {Very Deep Convolutional Networks for Large-Scale Image Recognition},
  booktitle = {International Conference on Learning Representations (ICLR)},
  year      = {2015}
}

@inproceedings{hog,
  author    = {Dalal, Navneet and Triggs, Bill},
  title     = {Histograms of Oriented Gradients for Human Detection},
  booktitle = {Proceedings of the IEEE Conference on Computer Vision and Pattern Recognition (CVPR)},
  pages     = {886--893},
  year      = {2005}
}

@inproceedings{alexnet,
  author    = {Krizhevsky, Alex and Sutskever, Ilya and Hinton, Geoffrey E.},
  title     = {ImageNet Classification with Deep Convolutional Neural Networks},
  booktitle = {Advances in Neural Information Processing Systems (NeurIPS)},
  year      = {2012}
}

@inproceedings{siddique2020,
  author    = {Siddique, Abubakar and Browne, Will N. and Grimshaw, Gina M.},
  title     = {Lateralized Learning for Robustness Against Adversarial Attacks in a Visual Classification System},
  booktitle = {Proceedings of the Genetic and Evolutionary Computation Conference (GECCO)},
  year      = {2020},
  pages     = {395--403},
  publisher = {ACM},
  doi       = {10.1145/3377930.3390164}
}

@article{mohammed2025,
  author  = {Mohammed, Thura J. and Xinying, Chew and Albahri, A. S. and Alnoor, Alhamzah and Wah, Khaw Khai},
  title   = {A Novel Trusted Framework for Orthopedic Disease Detection With Reliability Against Adversarial Attacks},
  journal = {IEEE Access},
  year    = {2025},
  volume  = {13},
  pages   = {160193--160220},
  doi     = {10.1109/ACCESS.2025.3609330}
}

@article{lal2021,
  author  = {Lal, Sheeba and Rehman, Saeed Ur and Shah, Jamal Hussain and Meraj, Talha and Rauf, Hafiz Tayyab and Damaševičius, Robertas and Mohammed, Mazin Abed and Abdulkareem, Karrar Hameed},
  title   = {Adversarial Attack and Defence through Adversarial Training and Feature Fusion for Diabetic Retinopathy Recognition},
  journal = {Sensors},
  year    = {2021},
  volume  = {21},
  number  = {11},
  pages   = {3922},
  doi     = {10.3390/s21113922}
}

\end{document}